\documentclass[letterpaper, 10 pt, conference]{ieeeconf}  

\IEEEoverridecommandlockouts                              

\usepackage{graphicx}
\usepackage{tabularx,booktabs}
\usepackage{hyperref}
\graphicspath{{figures/}{pictures/}{images/}{./}} 

\usepackage{array}
\newcolumntype{P}[1]{>{\centering\arraybackslash}p{#1}}

\title{\LARGE \bf
DecARt Leg: Design and Evaluation of a Novel Humanoid Robot Leg with Decoupled Actuation for Agile Locomotion
}

\author{Egor Davydenko$^{1}$, Andrei Volchenkov$^{1}$, Vladimir Gerasimov$^{1}$ and Roman Gorbachev$^{1}$
\thanks{$^{1}$Moscow Institute of Physics and Technology (MIPT)
        {\tt\small \{davydenko.ev, volchenkov.av, gerasimov.vn, gorbachev.ra\}@mipt.ru}}%
}

\begin{document}

\maketitle
\thispagestyle{empty}
\pagestyle{empty}

\begin{abstract}
In this paper, we propose a novel design of an electrically actuated robotic leg, called the DecARt (Decoupled Actuation Robot) Leg, aimed at performing agile locomotion. This design incorporates several new features, such as the use of a quasi-telescopic kinematic structure with rotational motors for decoupled actuation, a near-anthropomorphic leg appearance with a forward facing knee, and a novel multi-bar system for ankle torque transmission from motors placed above the knee. To analyze the agile locomotion capabilities of the design numerically, we propose a new descriptive metric, called the “Fastest Achievable Swing Time” (FAST), and perform a quantitative evaluation of the proposed design and compare it with other designs. Then we evaluate the performance of the DecARt Leg-based robot via extensive simulation and preliminary hardware experiments.
\end{abstract}

\section{Introduction}

In a search for a general-purpose humanoid robot, capable of effectively solving everyday tasks in practice, a trend can be observed, which can be named 'efficiency through simplicity'. In fact, most modern commercially available or open-source humanoid robots are representatives of a simple human-inspired leg morphology, often called a 'serial' or 'coupled' kinematic structure. \\

If we look at a range of modern humanoid robots like Unitree H1, Unitree G1, Fourier GR1, Booster H1, PnD Adam, Berkeley Humanoid, Westwood Robotics THEMIS and others, we can state that all of these robots have classic separate rotational hip, knee, and ankle actuators, placed in a serial kinematic chain. Some of these robots have a knee motor directly in the knee joint, some of them have a rod-and-lever transmission to actuate the knee joint from the motor placed in the hip. However, this configuration requires all actuators to be involved in performing the leg's swing motion, which is why this design can be named a coupled design \cite{coupledanddecoupled}. We believe that this tendency to prevalence of coupled (serial) anthropomorphic leg design arises from three major factors:

\begin{figure}[h]
        \centering
        \includegraphics[width=0.47\textwidth]{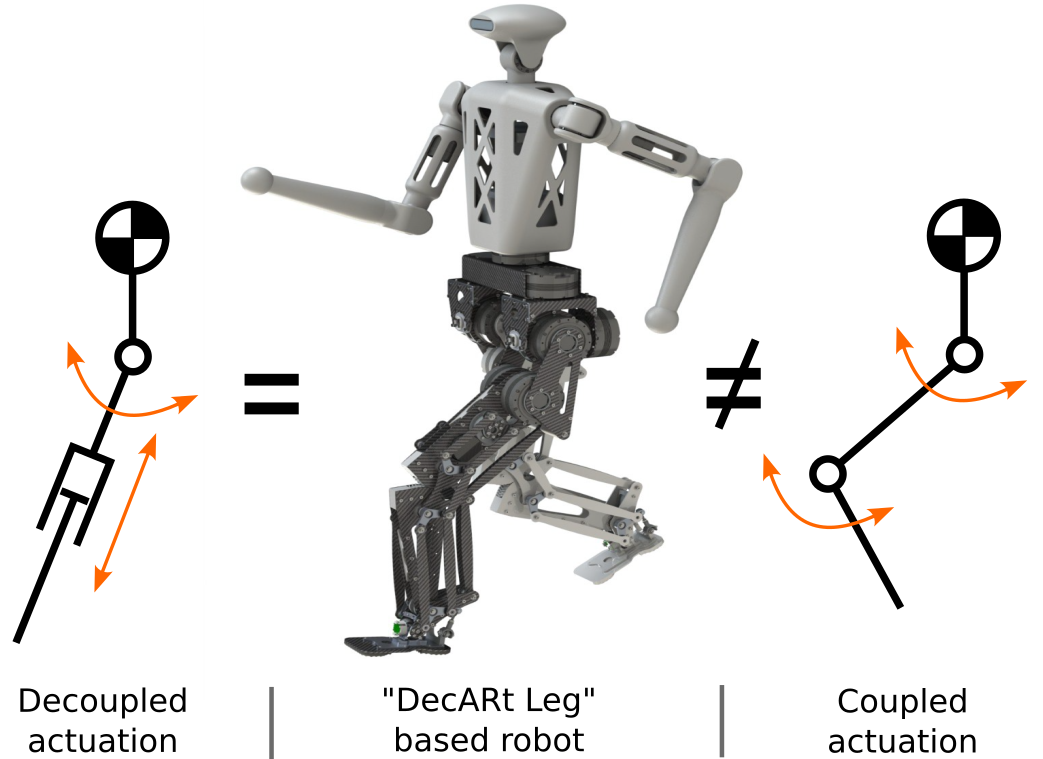}
        \centering
        \caption{The concept of DecARt Leg design: decoupled actuation, all motors above the knee, anthropomorphic appearance with a forward-facing knee.}
        \label{fig:servant_leg_intro}
\end{figure}

\begin{itemize}
        \item Intention to provide a user with a robot that can be simply modeled and controlled, not forcing the control engineer to deal with non-standard elements such as springs, tendons, or linear actuators integrated into complex closed kinematic chains.
        \item According to the 'uncanny valley' theory, a robot with a somewhat human-like appearance, but not too realistic, is perceived by humans as more friendly than one with a morphology that differs significantly from a human.
        \item It is expected that a robot with a morphology close to a human can effectively act in a human-centric environment - it can sit on the chair or car seat, directly use human furniture, etc.
\end{itemize}

But this tendency in the bipedal robot leg design can be treated as a limiting factor leading to a reduction of a robot's capabilities, especially in case if it can be proved that there are other humanoid robot leg designs still simple in control and visually resemble a human-like morphology, but more effective in some metrics than a classic coupled one.

In support of this statement, one can notice that there are some successful examples of a bipedal robot leg morphology which greatly differs from a human-like design, providing some advantages utilizing this non-classic kinematics. One remarkable example with proven efficiency is the Cassie/Digit robots leg design by Agility Robotics. 
In Cassie leg structure, the leg pitch actuator controls only the leg swing angle, not affecting the leg length, and the tarsus actuator in turn controls only the leg length, so we can treat Cassie leg morphology as a representative of a class of decoupled leg designs. 

Great research has been done on Cassie series robots that demonstrates the agile locomotion capabilities utilizing its decoupled design \cite{8814833}\cite{Reher2019DynamicWW}. At the time of writing, Cassie is the only robot capable of walking 5km on one battery or running a 100 m stride \cite{10160436}. But these advantages come with a price - the pantograph structure of a Cassie/Digit robot leg resembles a landbird more than a human, which in many cases one would like to avoid for a general-purpose humanoid robot.

So, the question arises: \textit{Is there any other possibility to create a design of a bipedal robot leg visually close to a human leg but still keeping the potential of a decoupled actuation}? In an attempt to answer this question, we present a novel design of a bipedal robot leg kinematics called the DecARt Leg.




The main contributions of the paper are as follows:
\begin{itemize}
        \item The novel leg design for an agile and energy-efficient bipedal locomotion, called the DecARt Leg, which facilitates a quasi-telescopic decoupled leg design with a novel multi-bar ankle actuation scheme.
        \item The numerical comparison of the capabilities of this design with other competitive designs.
        \item Formulation of a new metric, aimed at being descriptive and convenient to quantitatively evaluate agile locomotion capabilities of robotic legs.
        \item Evaluation of the proposed leg design in simulation.
        \item Preliminary evaluation of the proposed leg design in real hardware.
\end{itemize}

\section{Previous work and designs requirements}

The main goal of this work is to develop a new biped robot design that implements a decoupled actuation scheme, but still keeping the overall appearance of the robot close to a human-like morphology, and to demonstrate its agile locomotion capabilities. 

There are some remarkable robots that exemplify decoupled leg designs. These robots can be divided into two major categories: direct telescopic leg robots and quasi-telescopic leg robots.

The most straightforward implementation of the decoupled actuation scheme is a pure telescopic (prismatic) leg design. The main aim of existing robots with telescopic legs (Raibert Hopper \cite{Marc1984ExperimentsIB}, Meltran V \cite{MELTRANV}, ARL-monopod \cite{ARLmonopod}, SLIDER \cite{SLIDER}, LR04 \cite{LR04}) is to serve as research prototypes, capable of showcasing some interesting control methodologies and emphasizing novel principles in legged robot design, mostly using some form of a linear motors to perform the actuation. We can especially note the Raibert Hopper robot, which was the pioneer of a new age of dynamic legged locomotion. Despite their success, these robots clearly lack a human-like morphology. Another example of a decoupled robot is the PAL Kangaroo \cite{Kangaroo-roig:hal-03669855}, which demonstrates the ability to perform explosive motions such as jumping using complex multi-bar linkages and relying solely on linear actuators. In our research, we chose to use classic proprioceptive rotational actuators, same as those used in generic humanoid robots mentioned above, with the aim to preserve modeling and control simplicity.

The other interesting option to implement decoupled actuation on a real-world robot (and using rotational actuators) is a pantograph structure, which can be called “quasi-telescopic”. This structure was first introduced by ATRIAS robot \cite{ATRIAS} to emphasize the capabilities of a SLIP-based locomotion, and was later redesigned in the Cassie and Digit robot series by Agility Robotics. This leg design attracted significant attention in the robotic community, proving the high capabilities of its decoupled actuation scheme to achieve a robust and agile locomotion, especially being assisted with a proper placement of motors and series elastic elements \cite{ATRIASSLIP}. However, the pantograph structure of ATRIAS and Cassie/Digit legs is clearly not human-like. Also due to features arising from an impact mitigation capability analysis, the Cassie/Digit robot is preferred to walk with its knee (actually its enlarged heel) facing backward to maintain a low-impact posture configuration \cite{7139825}, further contributing to their bird-like appearance. As such, the challenge of achieving a human-like morphology while preserving a decoupled actuation scheme remains an open research problem.

The design requirements for agile locomotion focus on a robot’s ability to move its foot quickly enough to take long steps within a limited time, enabling high walking velocity and improved resistance to external disturbances. To meet these requirements, all actuators should be placed near the base of the leg to minimize swing inertia \cite{7139825}.

Consequently, the primary design requirements considered in this study are summarized as follows:
\begin{itemize}
    \item Decoupled actuation with a human-like leg appearance.
    \item Compliance with agile locomotion design principles, such as the placement of heavy motors near the root of the leg, while keeping the overall leg design compact and lightweight.
    \item Use of proprioceptive rotational actuators for ease of modeling and control.
    \end{itemize}
We follow these design principles in the next section.

\section{DecARt Leg design description}
We introduce a novel robotic leg design, named DecARt Leg, which uses the design requirements mentioned above in a high-performance structure designed for agile locomotion.

\begin{figure}[h]
        \centering
        \includegraphics[width=0.48\textwidth]{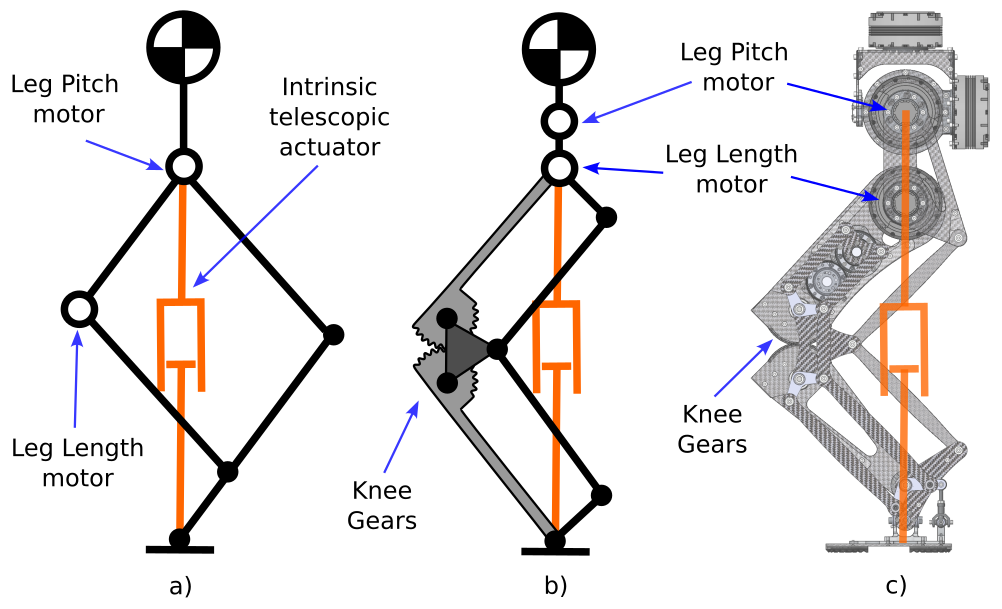}
        \centering
        \caption{Concept of the DecARt quasi-telescopic leg design: a) Pantograph-like design, b) DecARt Leg design, c) DecARt Leg CAD drawing and its intrinsic telescopic actuator.}
        \label{fig:servant_vs_pantograph_structure}
\end{figure}

\subsection{Design basis}
The basis of the DecARt Leg is a compact pantograph-like structure with a geared knee, which implements the decoupled design principle, allowing the leg actuation to function as a telescopic (prismatic) joint. The concept drawings of this design basis are presented in \autoref{fig:servant_vs_pantograph_structure}. The key feature that allows the leg kinematics to mimic a telescopic structure is a combination of a pair of passive gears and a 4-bar parallel structure. Due to this structure, the rotation of the leg length motor is directly transferred to a linear vertical motion of a foot, without involving the leg pitch motor. 

By incorporating an additional multi-bar linkage, all of the DecARt Leg motors are placed near the root of the leg, including those for the fully actuated 2-DoF ankle. 

We state that this leg design is capable of agile locomotion, which will be verified by quantitative experiments in Section \ref{fast_metric}. 

As seen from \autoref{fig:servant_leg_intro}, while intrinsically telescopic, the DecARt Leg can still be perceived to have a human-like appearance with a forward-facing knee, especially if the internal links are covered with appropriate aesthetic covers.

\begin{figure}[h]
        \centering
        \includegraphics[width=0.46\textwidth]{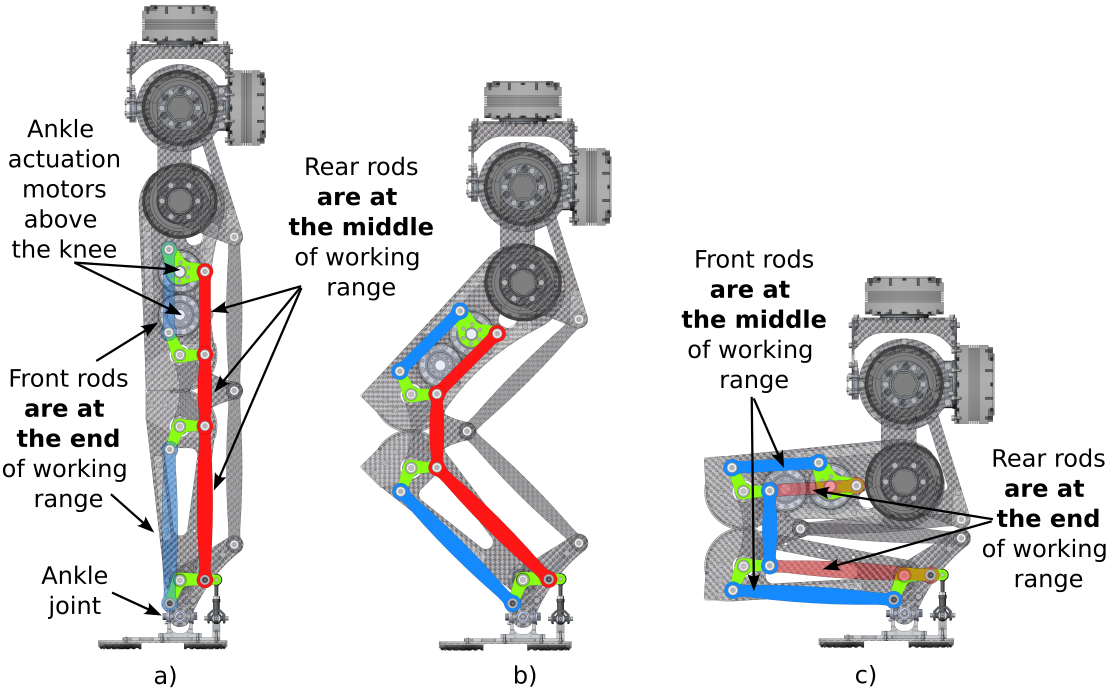}
        \centering
        \caption{DecARt Leg multi-bar ankle torque transmission structure: different rods are active during different periods of motion. Front rods (marked blue) and rear rods (marked red) are active during different poses of the leg: a) Fully stretched, b) Half-sitting, c) Fully crouched.} 
        \label{fig:ankle_rods_stand_mid_sitting}
\end{figure}

Despite its decoupled actuation design, the DecARt Leg structure provides some additional features, making it more flexible in real-world applications:
\begin{itemize}
\item The DecARt Leg can be fully compacted and stretched out (see \autoref{fig:ankle_rods_stand_mid_sitting}). This distinguishes the proposed design from a pantograph-based structure that cannot be fully stretched or fully compacted; otherwise its linkage structure will encounter a singularity issue.
\item The DecARt Leg is capable of altering the position of the ankle with a fully compacted knee (see \autoref{fig:ankle_rods_stand_mid_sitting}), while keeping the ability to maintain a dynamic balance in this pose. This is not the case for many modern bipedal robots (examples are Booster T1/K1, Fourier GR1) using bars and levers to control the ankle joint - these bars have a tendency to reach the boundaries of their working range in a fully compacted leg state.
\end{itemize}

\begin{figure}[h]
        \vspace{0.2cm}
        \centering
        \includegraphics[width=0.35\textwidth]{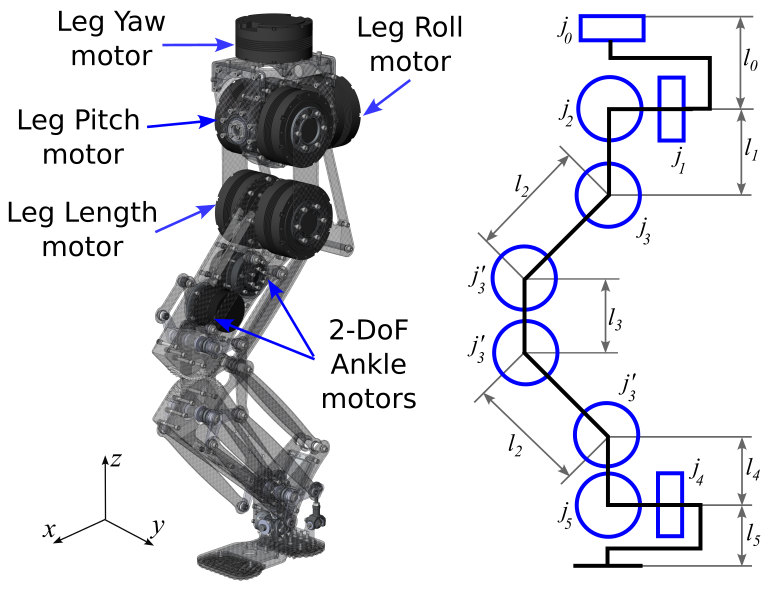}
        \centering
        \caption{DecARt Leg motors placement and kinematic structure. All motors are placed above the knee. Angles of joints $j'_3$ are equal to $j_3$ due to mechanical constraints of the quasi-telescopic design.}
        \label{fig:servant_ik}
\end{figure}

Also, due to the intrinsic presence of multiple geometric constraints, the inverse kinematics of the DecARt Leg can be derived analytically:


$$
j_0 = \psi \eqno{(1)}
$$
$$
j_1 = \arctan \big(\frac{y_r} {h}\big) \eqno{(2)}
$$
$$
j_2 = \arctan \big(-\frac{x_r} {h}\big) \eqno{(3)}
$$
$$
j_3 = -acos \Big(\big(\frac {h}{cos(j_1)cos(j_2)} - l_1 - l_3 - l_4\big) \frac {1}{2l_2}\Big) \eqno{(4)}
$$
$$
j_4 = \theta - j_1 \eqno{(5)}
$$
$$
j_5 = \phi - j_2 - j_3   \eqno{(6)}
$$
$$
x_r = x\cos(\psi)-y\sin(\psi);\  y_r = y\cos(\psi)+x\sin(\psi)  \eqno{(7)}
$$
where $\phi, \theta, \psi$ are the target foot roll, pitch, and yaw angles; $h$ is the target height from $j_2$ to $j_5$; $x, y$ - the target foot positions. Other values are according to \autoref{fig:servant_ik}.

The ability to derive a closed-form solution of inverse kinematics for the proposed leg structure, as well as other important quantities like frame Jacobians, simplifies the controller design and bridges the gap from the complex parallel structure to a simple serial leg design discussed above. 
To further support this statement, we note that we were able to directly model the proposed design in common modern simulators such as PyBullet and MuJoCo, and apply both classical and learning-based walking controllers to operate the DecARt leg-based robot in simulation and on real hardware. To simulate the closed kinematic chains of the DecARt design, the only additional requirement is joint mimic constraints for the joints labeled $j'_3$ in \autoref{fig:servant_ik}.

It is worth noting that, as seen from (6), the ankle pitch joint angle $j_5$ depends on both leg pitch actuator angle $j_2$ and leg length actuator angle $j_3$. This implies that the ankle pitch degree of freedom is not fully decoupled, similar to Cassie/Digit. Some ankle pitch actuation is required during leg length variation to fully mimic the ideal telescopic leg behavior. 

\section{Quantitative agile locomotion capabilities analysis}
\label{fast_metric}
One of the most straightforward metrics for evaluating a bipedal robot’s agile locomotion is its maximum achievable walking velocity. Although this metric is simple and descriptive, it is not directly suitable for comparing different bipedal leg designs. The first issue is that the fastest walking velocity largely reflects the capabilities of the walking controller used and its ability to stabilize the robot performing a fast gait. The second drawback is that this metric heavily depends on the robot's upper body characteristics, whereas we aim to choose a metric that solely evaluates only the leg design, not directly taking into account the robot's upper-body-related values like the total mass and inertia.

There are many metrics that are designed to evaluate the performance of a robotic leg. We can mention the Impact Mitigation Factor (IMF) \cite{IMF}, Centroidal Inertia Variation \cite{PyPNC}, Centroidal Inertia Isotropy \cite{TelloLeg}, Velocity Manipulability, Z-reduction ratio \cite{PinocchioComparitiveMetrics}, and Pratt Number \cite{PrattThesis}, among others. However, compared to the fastest achievable walking velocity, these metrics are less intuitive and descriptive.
A good candidate for a metric similar to the fastest walking velocity but suitable for leg design evaluation can be a leg's minimum swing time, still being the metric directly related to the fastest walking velocity \cite{PrattThesis}.

The use of minimum swing time as a quantitative measure of bipedal walking performance is not a new concept; for example, the Pratt Number  incorporated it into its calculation procedure \cite{PrattThesis}. However, to the best of the authors' knowledge, it has not yet been used as a standalone metric for directly comparing robotic leg designs across different scales.
One possible reason for this is that minimum swing time alone is a somewhat vague metric; its value depends heavily on the estimation method and other design variables. To use this value as a clear metric, important questions must be addressed — such as how to determine its value in a generic way suitable for various robotic leg designs and scales. In the next section, we address these questions and introduce a new metric, the "Fastest Achievable Swing Time".

\subsection{The Fastest Achievable Swing Time estimation}
The Fastest Achievable Swing Time (FAST) metric is estimated using a numerical evaluation of the robotic leg model. The metric allows direct comparison between robots with different leg designs and scales, without being biased by the robot's upper-body-related values such as total body mass and others. The core idea of this metric is not only to use minimum swing time for direct leg design comparisons, but also to rigorously define its estimation procedure, making it applicable to compare different robot scales and designs. Additionally, we provide an open-source package\footnote{\url{https://github.com/egordv/the_FAST_metric}} containing ready-to-use code for calculating the metric, to encourage other researchers to adopt it and reproduce our results.

To evaluate the metric, we perform an iterative numerical simulation of a leg swing motion, accounting for its full dynamics and actuation limits.

\begin{figure}[h]
        \vspace{0.2cm}
        \centering
        \includegraphics[width=0.48\textwidth]{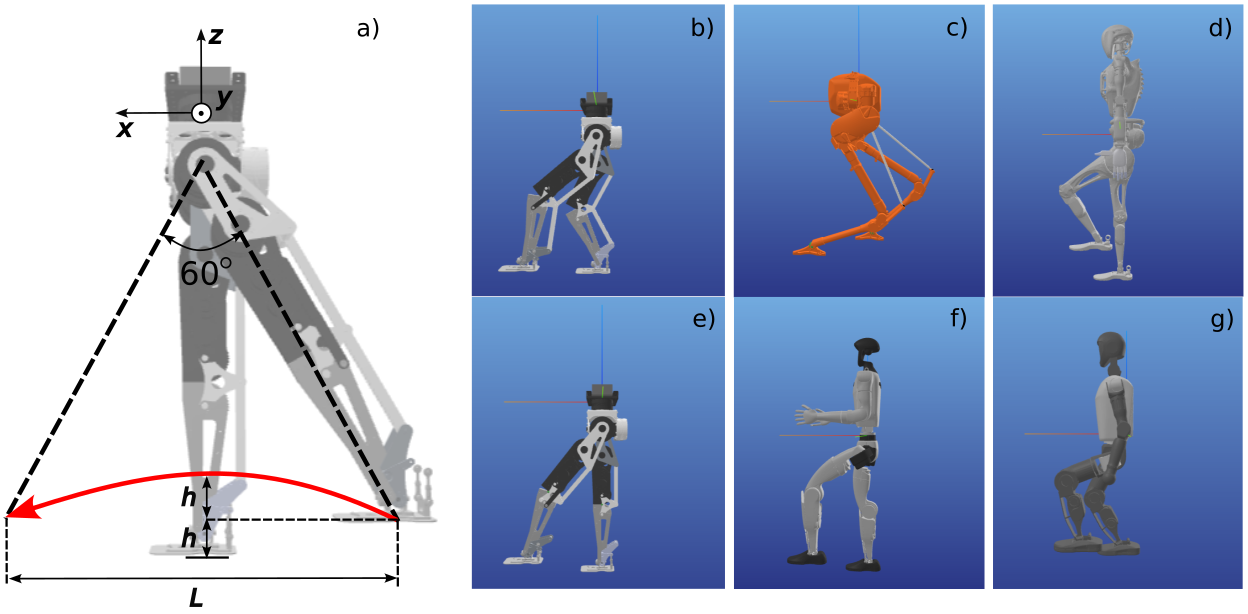}
        \centering
        \caption{ Fastest Achievable Swing Time (FAST) metric variables description (a), and screenshots of different robots doing the test: DecARt (b), Cassie (c), Fourier GR1T2 (d), DecARt-Serial emulation (e), Unitree G1 (f), and Booster T1 (g)}
        \label{fig:fastest_swing_test_desc}
\end{figure}

\begin{figure}[h]
        \vspace{0.2cm}
        \centering
        \includegraphics[width=0.48\textwidth]{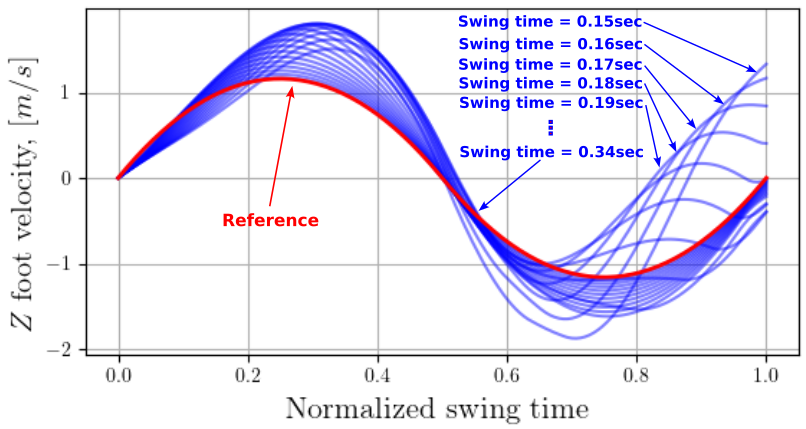}
        \centering
        \caption{ Iteratively reaching the required $Z$ foot velocity profile during the FAST metric evaluation.}
        \label{fig:fastest_swing_iterative}
\end{figure}

The procedure for estimating the metric is as follows:
\begin{itemize}
    \item A family of swing-foot trajectories is generated with the same swing length but different durations within a selected range. For this task, we use a cubic spline swing foot trajectory, treating it as a common choice in the legged robotics community. This solution is adopted from the ETH OCS2 package \cite{FARSHIDIAN20171463}. Although the optimal swing trajectory can differ for different leg designs, we chose the cubic spline for its generality and proven performance in practice.
\item Then, these trajectories are provided to the Qudratic Program (QP) optimization-based Inverse Dynamics (ID) controller as an operational space foot trajectory reference. We use a Task Space Inverse Dynamics Controller (TSID) \cite{adelprete:jnrh:2016} to calculate the required torques and accelerations of all joints to follow this swing trajectory with respect to the torque and velocity actuator limits and using the full robot's dynamic model provided via the URDF file. Once again, the controller used to perform a swing motion can differ for different robots, but QP-based controller is widely adopted solution, so we chose it for its generality and performance. 

\item After calculating the joint accelerations required for each trajectory, we numerically integrate them using the full robot dynamic model via the Pinocchio \cite{carpentier2019pinocchio} library.

\item Then we evaluate the operational space trajectories performed on the robot foot to comply with our requirements: During the entire swing period, the robot foot should reach the destination with a positional error less than 3\% of the swing length, and the velocity error during the entire trajectory should not be greater than $0.1m/s$.
The duration of the shortest swing trajectory that satisfies these requirements is selected as the final value of the metric.
\end{itemize}
The example of an evaluation of the generated trajectories iteratively reaching the required references is shown in \autoref{fig:fastest_swing_iterative}.

To be able to apply this metric for robots with different scales, we calculate the trajectory length of the swing leg and the height of the apex taking into account the actual size of the robot leg. The starting point of the swing trajectory is calculated for each robot by tilting its leg 30 degrees backward, and the final point of the swing trajectory is calculated by tilting the robot leg 30 degrees forward, resulting in a swing length of $L$ (see \autoref{fig:fastest_swing_test_desc}).
The height of the apex is selected to be equal to the absolute difference of the height of the foot during the initial tilt of the leg (value $h$ in \autoref{fig:fastest_swing_test_desc}). The robot base is assumed to be fixed so as not to alter the results of a leg metrics evaluation with the dynamics of the upper body. In addition, the excreted leg length is lowered by the factor of $0.98$ not to cause singularity issues in the QP controller.

Although the fastest achievable swing time is a descriptive metric itself, it is also the value that is the basis for several other important values. The first relevant value is the fastest walking velocity \cite{PrattThesis}. 
In fact, the duration of the swing is one of the main limiting factors when considering a fast walking gait. The corresponding theoretical walking velocity can be calculated directly from the swing duration and swing length values by simple division. Of course, during the stance phase, the leg motors are forced to exert torque much higher than during a swing phase to support the full robot's mass. But our aim is to develop a metric suitable to evaluate a sole leg design not depending on upper body weight, so we are not considering the stance phase in maximum walking velocity computation, that is why we call it a "theoretical walking velocity" (also ignoring some other factors, like the duration of double support phase if present). The degradation of this theoretical value when considering the complete robot structure with the upper body is demonstrated later in section \ref{simulation}.
Following the same logic, while mitigating external push, the robot is forced to quickly transfer the foot to the location required to catch the falling robot’s body, so the minimum achievable swing duration is directly related to the maximum possible push amplitude. 

It should also be noted that the value of the FAST metric obviously depends heavily on the limits of the leg's pitch motor, and can be over-fitted to enormously low value by placing an extremely powerful actuator to pitch the leg. In light of this, we emphasize that the FAST metric should not be used as the only metric to consider when developing a new robot. Its goal is to evaluate existing robotic leg structures that were designed with consideration of other important design variables, such as the robot's total mass, battery life and others, which effectively limit the actuator's size, weight of the mechanical linkages, and other factors that directly impact the final value of the FAST metric.


\subsection{Evaluation results}
We chose to evaluate the proposed metric and compare it to some other popular robots for which the URDF model is freely available. We chose to compare it with Cassie, treating this robot as an icon of dynamic bipedal locomotion capabilities using a decoupled leg design, Fourier Research GR1T2 robot as a reference for a coupled leg design featuring a lightweight leg with hip pitch and knee actuators placed near the hip joint, and Unitree G1 and Booster T1 robots as references for coupled leg design with knee actuators mounted directly in the knee joint. The estimated metric values are shown in Table \ref{table_fastest_swing_results}.

\begin{table}[h]
        \caption{The Fastest Achievable Swing Time Test Results}
        \label{table_fastest_swing_results}
        \begin{center}
        \begin{tabular}{|P{0.145\linewidth}|P{0.145\linewidth}|P{0.145\linewidth}|P{0.145\linewidth}|P{0.145\linewidth}|}
        \hline
        Robot model & Leg actuation design & Fastest achievable swing time, $s$& Corresp. swing length, $m$ & Theoretical walking velocity, $m/s$\\
        \hline
        Cassie & Decoupled & $0.24$ & $0.95$ & $3.94$\\
        \hline
        Fourier GR1T2 & Coupled & $0.24$ & $0.70$ & $2.92$\\
        \hline
        Unitree G1 & Coupled & $0.34$ & $0.64$ & $1.87$\\
        \hline        
        Booster T1 & Coupled & $0.33$ & $0.51$ & $1.55$\\
        \hline                
        DecARt & Decoupled & \textbf{0.17} & $0.71$ & \textbf{4.18}\\
        \hline                        
        DecARt-S & Coupled & $0.25$ & $0.71$ & $2.84$\\
        \hline                                
        \end{tabular}
        \end{center}
        \end{table}    
        
Analyzing this table, we observe that the DecARt leg design has a minimal achievable swing duration compared to the other models evaluated, which also corresponds to the highest theoretical walking velocity taking into account the scale of the robot. We believe that the main reason for such a short swing duration is that all DecARt Leg actuators are placed near the root of the leg, including ankle pitch and roll actuators, and its actuation structure is decoupled. 

To verify this statement and evaluate the importance of decoupled actuation, we modified the DecARt Leg URDF model into a coupled configuration by removing the associated closed kinematic chains and knee gears, and by locking the joint at the leg length motor location. 
After this modification, the two links above and below the leg length motor now form a rigid hip structure. As a result, the motor previously responsible for leg length actuation now becomes a hip-mounted actuator that remotely drives only the knee joint, similar to the configurations used in the MIT Humanoid and Fourier GR1. The overall motor placement and the total link masses were not changed by this modification. 

As a result, the leg design effectively becomes coupled (serial), with the leg pitch motor actuating only the thigh, and leg length motor actuating only the knee. This configuration is listed in the table as the “DecARt-S (serial)” model. We then evaluated this serial variant using the proposed metric, which showed a degradation of the minimum achievable swing duration value from 0.17 seconds to 0.25 seconds, under the same motor limits. 



In addition, this table clearly demonstrates that the placement of the actuator is an important design factor - all robots considered with knee actuators placed at the knee joint perform worse than those with actuators near the root of the leg.



\section{EVALUATION IN SIMULATION} \label{simulation}
To evaluate the proposed leg structure prior to its hardware implementation, we conducted extensive simulation-based evaluations. We created and validated the DecARt Leg URDF model, and extended it with a preliminary design of a torso and upper limbs. Then we put this model into a simulation environment using the PyBullet open-source simulation software. The total weight of the robot was approximately 35 kg. Then we tested the robot's ability to demonstrate an agile walk on flat and rough terrain, recover from pushes, climb stairs, open doors, and pick and carry a box and backpack. For a walking controller, we used an ALIPM-based \cite{9560821} controller, followed by a TSID Whole Body Controller (WBC) \cite{adelprete:jnrh:2016} with torque-controlled actuation. 
The simulation evaluation showed the following results: 
\begin{itemize}
    \item Maximum walking velocity with ALIPM control on flat terrain: $2.2 m/s$, on rough terrain: $0.8 m/s$ (with $5 cm$ terrain height variations)
    \item Maximum push force with ALIPM control: $95 N$ during $200 ms$ sagittal, $50 N$ during $200 ms$ lateral
    \item Maximum traversable stairs height: $10 cm$
    \item Maximum value of the additional torso weight: $8.5 kg$
    \item Maximum weight of the carried box (on exerted arms): $3.5 kg$
\end{itemize}
\begin{figure}[h]
        \centering
        \includegraphics[width=0.48\textwidth]{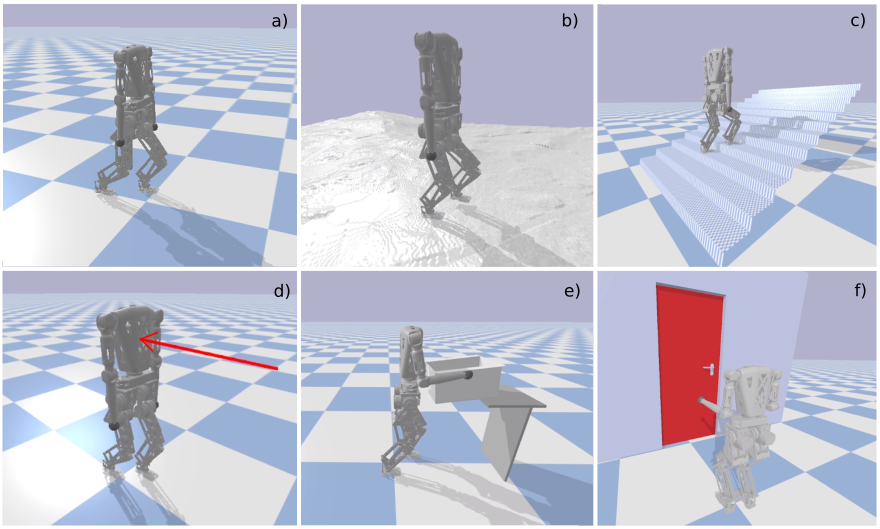}
        \centering
        \caption{Screenshots of different simulation scenarios to evaluate the DecARt Leg-based robot performance: Fast walk/walk with additional torso weight (a), Walk on rough terrain (b), Walk on stairs (c), Push recovery (d), Walking and picking a box from a table (e), Opening a door (f).}
        \label{fig:sim_samples}
\end{figure}
\autoref{fig:sim_samples} shows screenshots of the robot performing the above tasks. The results of the simulation demonstrate the promising locomotion capabilities of the proposed design, its ability to successfully perform real-world tasks, and the potential of further increasing the maximum walking velocity by the implementation of more advanced walking and running controllers.

\section{EVALUATION ON HARDWARE}
\begin{figure}[h]
        \vspace{0.2cm}
        \centering
        \includegraphics[width=0.48\textwidth]{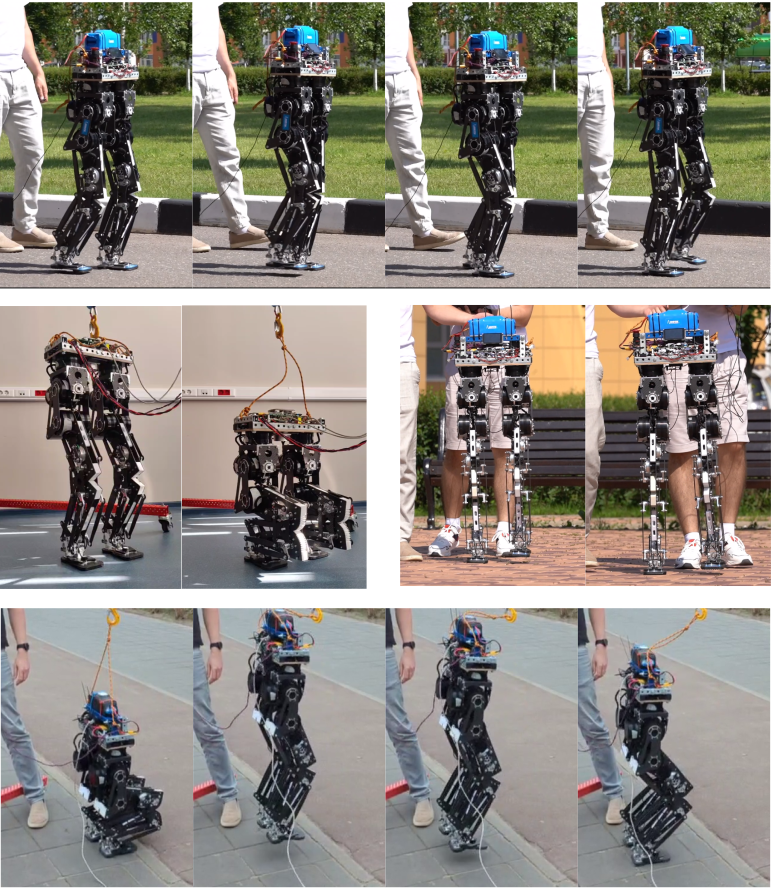}
        \centering
        \caption{Sample images of hardware evaluation: untethered walking (top), squatting (middle left), walking on semi-rough terrain (middle right), jumping (bottom)}
        \label{fig:hardware_samples}
\end{figure}

We conducted a preliminary hardware evaluation of the proposed leg design. At the time of writing, the two DecARt Leg units were manufactured and assembled into a bipedal platform with a rudimentary pelvis (see \autoref{fig:hardware_samples}). The design and manufacturing of the entire upper body of the robot is a topic of ongoing research. 
For hardware evaluation, we tested three control schemes: stiff position control, whole-body (WBC) torque control, and compliant PD control. 

In the dynamic walking hardware experiment, three walking controllers were preliminarily implemented and evaluated: an inverse kinematics-based controller derived from Equations (1–7), a QP-based WBC walking controller using the ALIPM framework \cite{9560821}, and a reinforcement learning controller based on \cite{RL_Walk}. The successful control and stabilization of the DecARt robot during a real-world walking task by all three controllers validates the simulation results and further support the goal of achieving simplicity in both modeling and control. Sample snapshots from the evaluation process are shown in \autoref{fig:hardware_samples}. To further evaluate the robot’s ability to perform explosive motions, we implemented a simple jumping controller based on a predefined trajectory, which also was successfully executed in practice.

Further parameter tuning, as well as a thorough quantitative hardware evaluation and comparison of the implemented walking controllers, including the verification of the design performance limits, are the subject of ongoing research.

\section{CONCLUSIONS}
In this work, we presented a new robotic leg design, the "DecARt Leg", which implements a decoupled actuation scheme inspired by pantograph-like mechanisms, using rotational proprioceptive actuators while maintaining an anthropomorphic appearance. To numerically compare the proposed kinematic structure with other structures, we proposed a new descriptive metric, called the “Fastest Achievable Swing Time” (FAST). 

Compared to the other coupled and decoupled leg designs considered, the DecARt leg's FAST metric value demonstrates strong potential for fast and agile locomotion. This result was supported by extensive simulation-based evaluation. The analysis of applying the proposed metric to various robot designs also revealed some meaningful insights, such as the direct performance difference between coupled and decoupled versions of the same base leg design, and the importance of actuator placement. The evaluation of the preliminary simulation model of the DecARt Leg-based robot across various tasks demonstrated significant potential for both locomotion and loco-manipulation.

Preliminary hardware evaluation of the DecARt Leg prototype confirmed its ability to perform locomotion and other motions like jumping in practice.
In our future work, we plan to implement the full DecARt Leg-based robot with torso and upper limbs in hardware and to conduct research on additional actuation elements, such as parallel and serial springs, to evaluate their impact on energy efficiency and locomotion performance.

\section*{ACKNOWLEDGMENT}    
We thank the members of the MIPT Laboratory of Wave Processes and Control Systems for their insightful help in completing this work. The authors thank Andrey Telichkin, Alexander Sukhanov, and Andrey Efremenko for their helpful support in setting up the robot hardware. We especially thank Azer Babaev for his motivation and support throughout this work.

\bibliographystyle{IEEEtran}
\bibliography{root}

\end{document}